\documentclass[11pt]{article}

\usepackage[final]{acl}

\usepackage{times}
\usepackage{latexsym}

\usepackage[T1]{fontenc}

\usepackage[utf8]{inputenc}

\usepackage{microtype}
\usepackage{tcolorbox} 

\usepackage{inconsolata}

\usepackage{graphicx}
\newcounter{todocounter}

\title{From Plausible to Actionable: A Position on LLM Self-Explanations}

\author{
  \textbf{Elize Herrewijnen\textsuperscript{1,2}\thanks{Corresponding Author:
  \href{mailto:email@domain}{e.herrewijnen@uu.nl}}},
\textbf{Benedetta Muscato\textsuperscript{3,4}},
\textbf{Gizem Gezici\textsuperscript{3}},
\textbf{Fosca Giannotti\textsuperscript{3}}
\\
\textsuperscript{1}University of Utrecht, Utrecht, Netherlands
\\
\textsuperscript{2}National Police Lab AI, Netherlands Police, Driebergen, Netherlands
\\
\textsuperscript{3}Scuola Normale Superiore, Pisa, Italy
\\
\textsuperscript{4}University of Pisa, Pisa, Italy
\\
}

\begin{document}
\maketitle
\begin{abstract}
Large Language Models (LLMs) can generate natural language explanations that rationalize their own decisions, a phenomenon commonly referred to as self-explanations.
Such explanations have emerged as a promising direction for explainable artificial intelligence (XAI), particularly for interpreting LLM behavior.
However, while self-explanations often appear plausible, whether they faithfully reflect a model's underlying reasoning process remains an open question.
In this opinion paper, we argue that self-explanations can be highly plausible, questionably faithful, and yet highly actionable.
From a traditional XAI perspective, we identify the limitations of standard evaluation protocols for LLM-generated self-explanations and propose practical guidelines for assessing their plausibility and faithfulness.
Moreover, we argue that evaluation should extend beyond these criteria to actionability, highlighting applications of LLM rationalization capabilities that support informed decision-making and appropriate action across diverse stakeholders.

\end{abstract}

\section{Introduction}
The emergence of Large Language Models (LLMs) has transformed the landscape of eXplainable Artificial Intelligence (XAI), creating new opportunities while introducing significant challenges \cite{zhao2024explainability}.
As LLMs are increasingly integrated into real-world applications, their explanations should accommodate the diverse goals, expertise, and technical backgrounds of stakeholders, thereby fostering informed trust \citep{calderon2025behalf}.

Traditional post-hoc XAI methods generate feature attributions, often using surrogate models such as LIME \citep{ribeiro2016should}. These approaches typically produce intuitive explanations, for example by highlighting the input tokens that influenced a prediction, making them relatively accessible to stakeholders. However, their computational cost limits scalability to modern LLMs with hundreds of billions of parameters \citep{jean2023cost}. 
Mechanistic interpretability offers a complementary approach by reverse-engineering the internal representations and computational mechanisms of models \citep{somvanshi2026bridging, bereska2024mechanistic, ranaldi2025survey}, but its complexity limits accessibility to stakeholders with advanced expertise \citep{wilde2025evaluating}.

LLM-generated self-explanations (hereafter, self-explanations) have emerged as a promising direction for bridging this gap.
Instead of relying on external explanation methods or probing internal mechanisms, LLMs can generate natural language explanations for their outputs \citep{fayyazEvaluatingHumanAlignment2024, paltaEverythingPlausibleInvestigating2026, ajwaniLLMGeneratedBlackboxExplanations2024} or produce intermediate reasoning through chain-of-thought (CoT) prompting \citep{wei2022chain}.
The natural-language format of self-explanations often creates a strong impression of transparency and alignment with human reasoning \citep{chen2026does, barez2025chain}, making them particularly appealing in high-stakes applications. 
For example, in healthcare, where LLMs are increasingly explored for clinical decision support and patient-facing applications, understandable explanations can help build trust among clinicians and patients \citep{afolabi2026faithful}.

Although self-explanations appear promising for explaining LLM decisions, whilst remaining accessible to many stakeholders, this does not necessarily imply that self-explanations align with human expectations or truly reflect the model's underlying decision-making process.
This distinction is captured by two central concepts in XAI: plausibility and faithfulness.
\emph{Plausibility} refers to how convincing or aligned with human reasoning an explanation appears \citep{jacovi2021aligning, paltaEverythingPlausibleInvestigating2026, lyu2024towards}, whereas \emph{Faithfulness} refers to the extent to which it accurately reflects the model's underlying decision process \citep{deyoung2020eraser, lyu2024towards}.

\paragraph{Contributions}
In this opinion paper, we argue that self-explanations can be highly plausible, questionably faithful, but potentially highly actionable. We substantiate this claim by identifying key limitations in existing protocols for evaluating the plausibility and faithfulness of self-explanations (\S\ref{sec:SE_as_XAI}). 
Given these limitations, we advocate shifting the focus towards the \emph{actionability} of self-explanations, that is their ability to help stakeholders make informed decisions and take concrete actions \citep{orgad2026interpretability}.
Accordingly, we outline a research agenda and possible ways to reposition self-explanations as communicative interfaces for stakeholders without technical expertise, as tools for supporting human decision-making and for facilitating deliberation from diverse perspectives (\S\ref{sec:alternative_applications}).
Finally, we provide preliminary directions for evaluating actionability and call for broadening evaluation beyond plausibility and faithfulness to account for the practical value of self-explanations.

\section{Self-Explanations: Highly Plausible, Questionably Faithful} \label{sec:SE_as_XAI}
Explanations generated by XAI methods are commonly evaluated along two key dimensions: plausibility and faithfulness \citep{doshi-velezRigorousScienceInterpretable2017, haseEvaluatingExplainableAI2020}.\footnote{We omit discussion of concepts such as simulatability (e.g., \citet{haseEvaluatingExplainableAI2020}) and monitoring (e.g., \citet{guan2025monitoringmonitorability}), as we believe they are rooted in the concepts of faithfulness and plausibility.} 
In the following, we examine the limitations of applying standard XAI evaluation protocols to self-explanations and propose practical guidelines for evaluating their plausibility and faithfulness.

\subsection{Plausibility}

Prior work has shown that self-explanations are often highly plausible and well aligned with human expectations \citep{paltaEverythingPlausibleInvestigating2026, ajwaniLLMGeneratedBlackboxExplanations2024, fanWhenAIPersuades2026}.
One factor contributing to this high level of plausibility is the sycophantic behavior of LLMs, whereby models tend to agree with or flatter stakeholders \citep{malmqvist2025sycophancy, wang2026truth}. As a result, stakeholders may be more likely to accept explanations that align with their expectations or existing beliefs.
In addition, \citet{palod2026evaluating, agarwalFaithfulnessVsPlausibility2024b} note that LLMs are optimized to generate coherent and helpful responses, which together with the accessibility of natural-language explanations, can increase their perceived interpretability \citep{madsenAreSelfexplanationsLarge2024a, kunz2024properties}.
The high plausibility of self-explanations can also be attributed to how plausibility is evaluated. In most studies, plausibility is assessed either through human ratings \citep{paltaEverythingPlausibleInvestigating2026, ajwaniLLMGeneratedBlackboxExplanations2024, fanWhenAIPersuades2026} or by comparing generated explanations with human explanations (e.g., \citet{fayyazEvaluatingHumanAlignment2024, deyoung2020eraser, bonaventuraExplanationAllYou}).
While these approaches capture important aspects of plausibility, we argue that they overlook two key elements. 

\paragraph{Plausibility Evaluation Ignores Stakeholder Expertise}
\citet{doshi-velezRigorousScienceInterpretable2017} propose explanation evaluation that includes both task experts (e.g., doctors) and lay users, where expert evaluations focus on explanation quality within the task context, and lay evaluations capture more general notions of explanation quality. This distinction is crucial for assessing plausibility, as a stakeholder's ability to judge an explanation depends on their expertise. For example, in a medical diagnosis task, a non-expert stakeholder may find an explanation with technical terminology convincing, whereas a doctor may recognize flawed reasoning or hallucinated content and judge it as less plausible. Thus, \textit{plausibility evaluation depends on the stakeholder’s expertise and background, and it should therefore be collected and assessed accordingly}.

\paragraph{Plausibility Evaluation Ignores Variation}
Generally, plausibility is measured by comparing a model's explanation with a human explanation, where closer alignment is interpreted as greater plausibility. However, this approach overlooks variation in how humans explain the same decision, particularly for ambiguous or subjective tasks. For example, in hate speech detection, annotators may interpret the same content differently and provide distinct labels and explanations based on their backgrounds and values. This variation naturally arises from instance difficulty, task ambiguity, and subjective interpretation \citep{plank2022problem, kulmizev2026label}. Consequently, relying on a single reference explanation risks collapsing diverse but valid perspectives into one standard. To better reflect human reasoning, plausibility evaluation should accommodate multiple valid explanations by aligning evaluation metrics with the task and validating them with domain experts \citep{muscato2026disagreeing}. Therefore, \textit{plausibility evaluation should embrace variation in ambiguous or subjective tasks, where multiple valid explanations may exist, rather than relying on a single reference explanation as the definitive standard.}

\subsection{Faithfulness}

A growing body of work has questioned whether self-explanations satisfy faithfulness requirements \citep{paulMakingReasoningMatter2024a, lanhamMeasuringFaithfulnessChainofThought2023, turpinLanguageModelsDonta, lewis2025analysing, madsenAreSelfexplanationsLarge2024a, colegadoFaithfulPlausibleCounterfactual2026}. 
We argue that self-explanations are questionably faithful for three main reasons.

First, LLMs lack access to the internal processes that generates their outputs.
As argued by \citet{sarkar2024large}, self-explanations are generated through the same process as prediction, without access to meta-information about the prediction process itself. Therefore, it is unlikely that self-explanations reflect genuine introspective analysis.
Second, faithful explanations require separating decision-making from explanation generation. However, prompting models to explain their predictions may influence the decision process itself, as suggested by the observed impact of self-explanations on prediction accuracy \citep{li2023towards}.
Third, LLMs may provide coherent but misleading rationales due to sycophantic behavior, hallucinations, or biases \citep{feng2026good}, raising doubts about whether self-explanations reflect the true causes of their decisions.
As a result, when models produce incorrect or hallucinated outputs \citep{lin2025investigating}, their explanations may still present coherent but misleading justifications, raising concerns about whether the reported rationale truly corresponds to the underlying cause of the decision.

Despite the above concerns, several studies have investigated the faithfulness of self-explanations \citep{paulMakingReasoningMatter2024a, huangCanLargeLanguage2023, maynePositiveCaseFaithfulness2026, turpinLanguageModelsDonta, cornish2025examining, chen2025reasoning, chua2025deepseek}. 
While we encourage investigating the faithfulness of self-explanations, \emph{we argue that existing faithfulness evaluation methods are not suitable for self-explanations}. Many of these methods are perturbation-based, 
where model inputs are perturbed (i.e., removing explanation words) and model outputs (i.e., predictions) for these perturbed inputs are compared \citep{ribeiro2016should, huangCanLargeLanguage2023, fayyazEvaluatingHumanAlignment2024, deyoung2020eraser, parcalabescu2024measuring, atanasova-etal-2023-faithfulness}. These methods rest on three assumptions identified by \citet{jacoviFaithfullyInterpretableNLP2020}: the model assumption, the prediction assumption, and the linearity assumption. When it comes to LLMs and self-explanations, these assumptions do not hold for three main reasons.

\paragraph{Faithfulness Evaluation Overlooks Nondeterminism in LLMs}
The first assumption identified by \citet{jacoviFaithfullyInterpretableNLP2020} is the \emph{model assumption}, which states that ‘two models will make the same predictions if and only if they use the same reasoning process.’
Nondeterminism complicates the practical applicability of the model assumption for LLMs, because identical model--input pairs can yield different outputs across runs, and their self-explanations can vary as well \citep{astekinExploratoryStudyHow2024, atilNonDeterminismDeterministicLLM}.
Moreover, \citet{bogaert2025consolidating} provide empirical evidence that models with comparable predictive performance may nonetheless produce substantially different explanations for their decisions. 
Therefore, \textit{faithfulness evaluation should account for the nondeterministic behavior of LLMs, rather than treating single-run outputs as definitive evidence of faithfulness.}

%

\paragraph{Faithfulness Evaluation Overlooks Prompt Sensitivity in LLMs}
The second assumption, the \emph{prediction assumption}, states that ‘on similar inputs, the model makes similar decisions if and only if its reasoning is similar.’ 
However, for LLMs, the notion of a ‘similar input’ is often narrow: models can be highly sensitive to subtle surface-form changes such as spacing, capitalization, and punctuation \citep{he2024does, zhuo2024prosa, loya2023exploring, errica2025did, gan2024reasoning}.
As a result, prompts that are semantically equivalent or nearly identical to humans may elicit different predictions. For instance, removing a comma from the input can unexpectedly change the model’s output. 

This sensitivity complicates the use of input similarity as a stable basis for faithfulness evaluation. Existing faithfulness evaluations often incorporate input perturbations implicitly, but they rarely treat prompt sensitivity itself as a separate variable to measure systematically.
This is also true for prompt modifications that appear unrelated to the decision itself. For example, asking the LLM to provide explanations \citep{huangCanLargeLanguage2023, mcmillan2025towards} or 
adding `Please answer this question' to the prompt \citep{chua2025deepseek} can negatively affect predictive performance.
Therefore, \textit{faithfulness evaluation should not rely on an assumed (dis)similarity between inputs, but should account for prompt sensitivity in LLMs as a distinct source of variation.}

\paragraph{Perturbation-Based Faithfulness Evaluation Is Unreliable for LLMs}
Finally, the 
\emph{linearity assumption} poses that ‘certain parts of the input are more important to the model reasoning than others. Moreover, the contributions of different parts of the input are independent from each other’ \citep{jacoviFaithfullyInterpretableNLP2020}.
Relying on this assumption, faithfulness metrics commonly remove or mask parts of the input and assess the change in the model's predictions \citep{deyoung2020eraser, ribeiro2016should}. This approach is not suitable for LLMs, because an LLM may disregard the task input and instead rely solely on its internal knowledge to complete the task.
For example, LLMs may continue to predict a sentiment even after all sentiment-bearing words have been removed from the input, and have been shown to solve ‘unanswerable’ tasks by producing predictions for inputs that lack the information required to determine the correct answer \citep{fayyazEvaluatingHumanAlignment2024, ming2025faitheval, kirichenko2025abstention}.
From the perspective of faithfulness evaluation, such behavior can itself be interpreted as evidence of unfaithful explanations: if an explanation attributes a prediction to specific input features, yet the prediction remains unchanged when those features are masked, the explanation is not faithfully reflecting the true basis of the model’s decision.

Moreover, perturbing inputs may induce unwarranted model behaviour (e.g. refusal \citep{huangCanLargeLanguage2023}), especially when the perturbation increases the unnaturalness of the text. This also applies to perturbation techniques that rely on synthetic inputs, such as counterfactuals  \citep{colegadoFaithfulPlausibleCounterfactual2026,  madsenAreSelfexplanationsLarge2024a}. Such synthetic inputs may contain hidden patterns that LLMs can exploit, e.g., self-preference biases \citep{panickssery2024LLrecognize, wataoka2024self}, potentially allowing the model to game faithfulness tests rather than reveal its true reasoning.
Thus, \textit{faithfulness evaluation should account for two key concerns: (i) the model's reliance on parametric knowledge over the task input, and (ii) the risk that perturbed or synthetic inputs introduce exploitable artifacts rather than meaningful signal.}

\subsection{High Plausibility Can Mask Unfaithfulness} 
Explanations that are highly plausible, but questionably faithful can be misleading. Importantly, an LLM may produce an incorrect prediction while providing an explanation that faithfully reflects the reasoning underlying that prediction. Conversely, an explanation may appear convincing while failing to faithfully reflect the factors that actually influenced the model’s prediction. This latter case, in which plausibility may mask a lack of faithfulness, raises concerns about the interpretation and use of LLM self-explanations.
\citet{paltaEverythingPlausibleInvestigating2026} and \citet{fanWhenAIPersuades2026} show that humans often find LLM self-explanations persuasive even when the model’s prediction is incorrect. \citet{ajwaniLLMGeneratedBlackboxExplanations2024} refer to this phenomenon as \emph{adversarial helpfulness}, where explanations make incorrect predictions appear correct. Such effects may reduce users’ ability to detect erroneous model outputs \citep{fanWhenAIPersuades2026}, potentially leading to unwarranted trust in the LLM's capabilities. This trust may foster automation bias, i.e., the tendency of users to over-rely on automated recommendations \citep{romeo2026exploring}. Such over-reliance is particularly concerning in high-stakes domain \citep{jacovi2021formalizing}.

\section{Towards Actionable Self-Explanations}
\label{sec:alternative_applications}
In the previous section, we discussed the challenges of evaluating the plausibility and faithfulness of self-explanations. Although such self-explanations can appear highly plausible, prior work has shown that they are not always faithful, as they may not accurately reflect the model’s true decision-making process (\S\ref{sec:SE_as_XAI}). Given the limitations of current faithfulness evaluation methods, we suggest that their value may extend beyond strictly reflecting this process. 
Thus, we advocate viewing self-explanations through the lens of \emph{actionability}: their ability to help diverse stakeholders make better-informed decisions and take effective actions \citep{orgad2026interpretability}.

Crucially, we view the benefits of actionable self-explanations as downstream of faithfulness: an unfaithful explanation cannot be safely acted upon. At the same time, faithfulness alone does not guarantee actionability. An explanation may faithfully track a model’s decision process yet remain too technical, opaque, or misaligned with users’ tasks to support concrete decisions. \textit{Our position is therefore that the community should move beyond evaluating faithfulness in isolation and explicitly design and evaluate self-explanations that are both faithful and actionable for end users}. In this view, actionability is a necessary complement to faithfulness, not a substitute for it.

We treat self-explanations as outputs of an LLM's rationalization capability, i.e., its ability to generate justifications for decisions. From this perspective, we highlight three practical applications of LLM rationalization: (i) serving as a communicative interface that makes XAI more accessible to non-expert stakeholders; (ii) supporting human decision-making; (iii) facilitating deliberation by surfacing diverse perspectives.

\paragraph{Making XAI Accessible to Non-Expert Stakeholders}
A key challenge in the practical adoption of XAI is that the most faithful explanations are often the least accessible to non-expert stakeholders \cite{colin2022cannot}.
Self-explanations can bridge this gap by translating faithful but complex explanations into natural language rationales that are easier to understand and act upon \citep{serafim2024exploring, serafim2025mainle, mermigkis2026twostagellmframeworkaccessible}. They can also adapt explanations to different stakeholders by adjusting terminology, formulation, and level of detail to the expertise and information needs of stakeholders \citep{albert2024user}.
Nonetheless, as also noted by \citet{serafim2025mainle}, LLM-generated narratives can distort otherwise faithful XAI outputs by over-interpreting, omitting, or hallucinating causal claims. This risk is especially acute in high-stakes settings, where a fluent but unfaithful explanation may be more misleading than no explanation at all.
We therefore argue that any pipeline that verbalizes XAI signals through LLMs should incorporate lightweight safeguards such as constrained prompting, structured output schemas, and a verification step to reduce the likelihood of unfaithful narrations.
\textit{Self-explanations should thus not be treated as standalone explanations, but as communicative interfaces that render faithful XAI methods more accessible}.

Self-explanations could also communicate uncertainty and potential hallucinations in LLM outputs \citep{al2025llm, banerjee2025llms}. Although a growing body of work has focused on detecting hallucinations in LLMs \citep{farquhar2024detecting, sriramanan2024llm}, self-explanations can complement these methods by conveying their signals in natural language, helping users better assess the reliability of model outputs.
This perspective aligns with \citet{ulmer2025anthropomimetic}, who argue that LLMs should communicate uncertainty in a human-like manner.

\paragraph{LLMs in Support of Human Decision-Making}

In high-stakes domains, AI systems should be designed to support human judgment instead of replacing it \citep{kostick2022ai,lazaros2026human,panigutti2023role}. Rather than treating self-explanations as faithful accounts of model reasoning, \emph{we argue they should be viewed as arguments that human decision-makers can critically evaluate}. Their value lies not in revealing how a model reached a conclusion, but in surfacing assumptions, uncertainties, and alternative perspectives that help decision-makers critically evaluate the generated arguments and form their own judgement \citep{sarkar2024ai}.
This perspective can be operationalized through safety protocols (e.g., guardrails \citep{jalan2026survey}), that constrain LLMs to generate arguments rather than final decisions. For example, in the medical domain, the LLM could present the clinical evidence that supports or challenges candidate diagnoses, leaving the final decision to the doctor.

While our discussion mainly focuses on LLMs supporting human decision-making, our perspective also extends to autonomous or semi-autonomous workflows. When LLMs act as decision-makers, self-explanations can serve as actionable artifacts for post-hoc auditing, debugging, and accountability, enabling developers, regulators, and affected users to understand and contest automated decisions. In such settings, the safeguards we advocate (e.g., constrained prompting, verification layers) become even more critical given the lack of real-time human oversight. We acknowledge, however, that in fully autonomous, high-stakes deployments, self-explanations alone are insufficient and must be complemented by stronger reliability guarantees and monitoring mechanisms.

Nevertheless, we recognize that in low-stakes scenarios, end-to-end LLM reasoning and judgment may be acceptable, or even preferable.

\paragraph{LLMs as Advocates for Deliberation}
Viewing self-explanations as a means of supporting human decision-making, we argue that such explanations can foster deliberation through diverse perspectives \citep{vijjini2024socialgaze}. Under this framing, variation is essential: different LLMs can surface complementary arguments and counterarguments that enrich the decision process (e.g., \citep{li-etal-2024-llms-speak}). We call this paradigm \emph{LLMs as advocates, where models provide diverse viewpoints while humans remain the final decision-makers}. By expanding the range of considerations available, these systems have the potential to enhance human judgment and foster reflective deliberation.

\paragraph{Future Insights for Evaluating Actionability}
While actionability may be more difficult to measure than plausibility or faithfulness, it is not intractable \cite{alotaibi2026explain}. For each of our three practical applications, we outline initial evaluation directions:
\begin{itemize}
    \item \textit{Making XAI Accessible to Non-Experts}. Evaluate whether self-explanations help non-experts correctly interpret XAI outputs through a participatory approach, in which users assess the quality and usefulness of explanations for their intended tasks, identify uncertainty and potential errors, and determine when to trust, override, or seek human review.
    \item  \textit{LLM-Supported Human Decision-Making}. Use decision-impact experiments to assess whether self-explanations improve task performance (accuracy, calibration, time, workload) and decision quality (e.g., fewer harmful overrides, better risk-sensitive choices) versus no- or baseline-explanation conditions (e.g., \citep{Haag02012026}).
    \item \textit{LLMs as Deliberation Advocates}. Assess whether multi-perspective self-explanations broaden considerations, reduce premature convergence, and improve group decision quality using deliberation transcripts and metrics of argument diversity and outcomes.

\end{itemize}

\section{Conclusion}

In this opinion paper, we have highlighted the limitations of existing XAI evaluation frameworks for measuring the plausibility and faithfulness of LLM-generated self-explanations.
Nevertheless, we argue that self-explanations derive their value from their actionability: their ability to support informed decision-making and enable stakeholders with diverse goals, expertise, and backgrounds to take appropriate actions.
We position self-explanations not as literal accounts of a model's internal reasoning, but as communicative interfaces for non-experts, decision-support tools, and facilitators of deliberation by presenting diverse perspectives.
Overall, we advocate moving beyond treating plausibility and faithfulness as the primary objectives of self-explanations, and instead emphasizing their actionability in enabling informed decision-making across diverse stakeholders.

\section*{Limitations} 
While our opinion paper provides insights into how existing plausibility and faithfulness evaluation protocols could be challenged and reframed towards more actionable self-explanations, we do not provide a practical implementation or empirical evaluation of the proposed revisions.
Moreover, our aim is not to introduce new evaluation metrics, but rather to offer a fresh perspective for the XAI research agenda by highlighting the importance of developing and evaluating explanations that are more actionable for diverse stakeholders.

\section*{Acknowledgments}
This work has been supported by the National Police Lab AI (NPAI), the European Union under ERC-2018-ADG
GA 834756 (XAI), the Partnership Extended PE00000013 - “FAIR - Future Artificial Intelligence Research” - Spoke 1 “Human-centered AI” and the European Union under Grant Agreement no. 101120763 “TANGO”. 
We thank the members of the NLP and Society (Utrecht) and DHAI (Pisa) labs for their insightful feedback on earlier drafts of this paper. We specifically appreciate the suggestions of Dong Nguyen, Albert Gatt, and Marzio Di Vece.
Finally, we thank the anonymous reviewers for their helpful suggestions.

\bibliography{custom}

\end{document}